\documentclass{article}
\usepackage{ijcai21}

\usepackage{times}
\usepackage{soul}
\usepackage{url}
\usepackage[hidelinks]{hyperref}
\usepackage[utf8]{inputenc}
\usepackage[small]{caption}
\usepackage{graphicx}
\usepackage{amsmath}
\usepackage{amsthm}
\usepackage{booktabs}
\usepackage{algorithm}
\usepackage{algorithmic}
\usepackage{makecell}
\usepackage{multirow}

\usepackage{epsfig}
\usepackage{amssymb}
\title{RR-Net: Injecting Interactive Semantics in Human-Object Interaction Detection}

\author{
Dongming~Yang$^1$\and
Yuexian~Zou$^{1,2}$\footnote{Contact Author}\and
Can~Zhang$^{1}$\and
Meng~Cao$^{1}$\And
Jie~Chen$^{1,2}$\\
\affiliations
$^1$School of ECE, Peking University, Shenzhen, China, 518055\\
$^2$Peng Cheng Laboratory, Shenzhen, China, 518055\\
\emails
\{yangdongming, zouyx, zhangcan, mengcao\}@pku.edu.cn,
chenj@pcl.ac.cn
}
\begin{document}

\maketitle

\begin{abstract}
%via inferring fine-grained triplets of $\langle$ human, verb, object $\rangle$
Human-Object Interaction (HOI) detection devotes to learn how humans interact with surrounding objects. Latest end-to-end HOI detectors are short of relation reasoning, which leads to inability to learn HOI-specific interactive semantics for predictions. In this paper, we therefore propose novel relation reasoning for HOI detection. We first present a progressive Relation-aware Frame, which brings a new structure and parameter sharing pattern for interaction inference.
%Upon the frame, an Interaction Intensifier Module is then designed, where interactive semantics from humans can be exploited and passed to objects to intensify interactions.
%Furthermore, a Correlation Parsing Module is proposed, which promotes predictions by integrating interactive correlations among humans, objects and interactions. 
Upon the frame, an Interaction Intensifier Module and a Correlation Parsing Module are carefully designed, where: a) interactive semantics from humans can be exploited and passed to objects to intensify interactions, b) interactive correlations among humans, objects and interactions are integrated to promote predictions.
Based on modules above, we construct an end-to-end trainable framework named Relation Reasoning Network (abbr. RR-Net). 
Extensive experiments show that our proposed RR-Net sets a new state-of-the-art on both V-COCO and HICO-DET benchmarks and improves the baseline about 5.5\% and 9.8\% relatively, validating that this first effort in exploring relation reasoning and integrating interactive semantics has brought obvious improvement for end-to-end HOI detection.
%We will release our code upon publication.
\end{abstract}

\section{Introduction}

Fine-grained understanding of visual contents is one of the fundamental problems in computer vision. The task of Human-Object Interaction (HOI) detection aims to localize and infer triplets of $\langle$ human, verb, object $\rangle$ from a still image. Beyond comprehending instances, e.g., object detection \cite{ren2017faster,YANG201920}, human pose estimation \cite{fang2017rmpe} and action recognition \cite{sharma2015action}, detecting HOIs requires a deeper understanding of visual semantics to depict complex and finer-grained relationships between $\langle$ human, object $\rangle$ pairs. 
%%%HOI detection presents more difficulties compared with traditional action recognition \cite{sharma2015action}, e.g., an individual can simultaneously take multiple interactions with surrounding objects. 
%%%Besides, associating ever-changing roles with various objects lead to finer-grained and diverse samples of interactions. 
Therefore, HOI detection is challenging and usually needs to mine high-level and HOI-specific representations in solutions.
\begin{figure}[t]
\begin{center}
\includegraphics[width=1\linewidth]{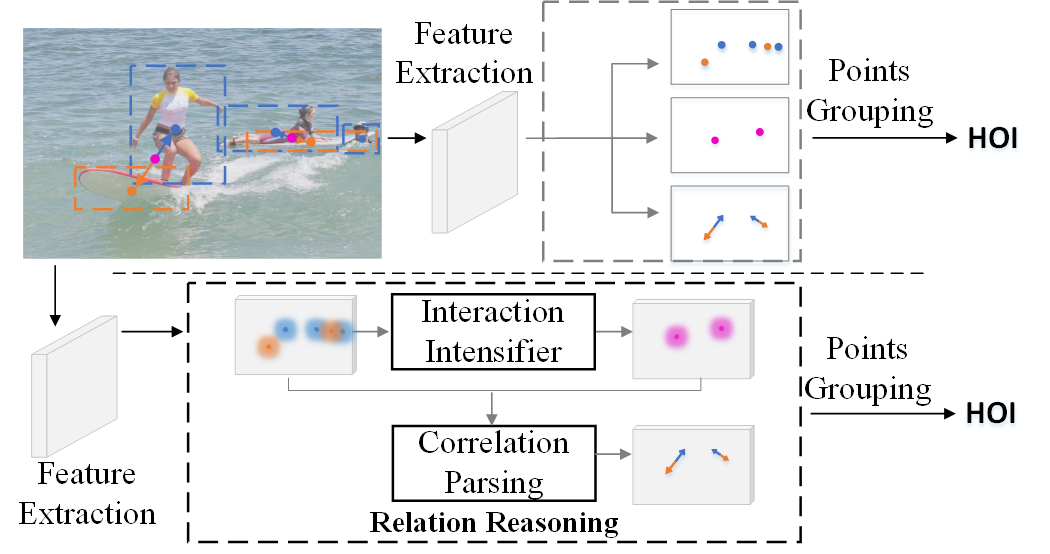}
\end{center}
\vspace{-0.5cm}
   \caption{Visualization of how our method facilitates HOI detection. The baseline framework (above) lacks learning relative information during inference, while we present a framework (bottom) that performs relation reasoning to learn HOI-specific interactive semantics.}
\label{FIG:1}
\vspace{-0.5cm}
\end{figure}

The conventional two-stage HOI detection methods \cite{gkioxari2018detecting,gao2018ican:,xu2019interact} first obtain human and object region proposals from a pre-trained object detector \cite{ren2017faster} and then generate $\langle$ human, object $\rangle$ pairs to make interaction classifications. The related techniques of HOI feature learning are all region-based where instance (e.g., human and object) attention features \cite{gkioxari2018detecting,gao2018ican:,xu2019interact}, spatial configuration \cite{gao2018ican:,No-Frills_2019_ICCV,VSGNet_2020_CVPR} and human pose \cite{wan2019pose,zhou2019relation} are exploited. Inspired by the point-based object detectors \cite{law2018cornernet,zhou2019objects}, as illustrated in Figure~\ref{FIG:1} (above), latest end-to-end HOI detectors pose HOI detection as a keypoint detection and grouping problem, which effectively improve the speed of HOI detection. 

%%Although previous feature learning techniques generally benefit HOI representations, they bring several drawbacks: 1) Firstly, dominant methods focus on region-based feature enhancement but lack relation reasoning, however, comprehending HOI-specific interactive semantics from relation dependency among humans, objects and interactions is significant for HOI inference. 2) Secondly, region-based feature learning is complicated, and employing human pose or body-part features brings additional workload. Besides, previous region-based feature enhancements are no longer applicable to end-to-end detectors due to these end-to-end frameworks do not provide proposed instance regions during inference. 
Previous region-based feature enhancements are hardly applicable to end-to-end detectors due to these end-to-end frameworks do not provide proposed instance regions during inference.
Besides, although previous feature learning techniques generally benefit HOI representations, they bring several drawbacks: 1) Firstly, dominant methods focus on region-based feature enhancement but lack relation reasoning, however, comprehending HOI-specific interactive semantics from relation dependency among humans, objects and interactions is significant for HOI inference. 2) Secondly, region-based feature learning is complicated, and employing human pose or body-part features brings additional workload. 
Therefore, studying relation reasoning to integrate region-independent interactive semantics has become the key to improve end-to-end HOI detection.

In this work, as illustrated in Figure~\ref{FIG:1} (bottom), we propose relation reasoning for HOI detection, which captures and transmits HOI-specific interactive semantics among humans, objects and interactions where no human and object proposals are needed.
The contributions are summarized as follows:
\begin{itemize}
    \item A progressive Relation-aware Frame is first presented to build a coherent structure and parameter sharing pattern for interaction inference.
    %%%, which imitates the human visual mechanism of recognizing HOI by comprehending visual instances and interactions coherently.
    \item Upon the frame, an Interaction Intensifier Module (abbr. IIM) is proposed, where interactive semantics from humans can be exploited and easily passed to objects. 
    %%%Therefore, the objects with high relevance with humans will be relatively valued to intensify interactions.
    \item A Correlation Parsing Module (abbr. CPM) is designed subsequently, in which interactive correlations among visual targets can be exploited and integrated. 
    %%%Such a module helps group HOI triplets without instance regions.
    \item We offer a new end-to-end framework named Relation Reasoning Network (abbr. RR-Net), which is the first to carry out region-independent relation reasoning for HOI detection. 
\end{itemize}
We perform extensive experiments on two public benchmarks (i.e., V-COCO \cite{yatskar2016situation} and HICO-DET \cite{chao2018learning} datasets). Our method with relation reasoning provides obvious performance gain compared with the baseline and outperforms the state-of-the-art methods by a sizable margin. Detailed ablation studies of our framework are also provided to facilitate future research.

\section{Related Work}

\subsection{Frameworks in HOI Detection}
%Beyond detecting instances from images, visual semantic role labeling \cite{yatskar2016situation} is introduced to learn interactions between humans and objects. 
Since human and object instances are basic elements in detecting HOIs, conventional HOI detectors are mostly based on object detection frameworks. InteractNet \cite{gkioxari2018detecting} extends the object detector Faster R-CNN \cite{ren2017faster} with an additional branch for HOI detection. iCAN \cite{gao2018ican:} then introduces a three-branch architecture with one branch each for a human candidate, an object candidate, and an interaction pattern, which has become the template framework for most later HOI detectors. TIN \cite{li2019transferable} further proposes a multiple network to perform feature extraction, interactiveness discrimination and HOI classification, respectively. All above frameworks are two-stage and region-based, where region proposals obtained from pre-trained object detectors are indispensable. 
%%%Subsequent researches \cite{VSGNet_2020_CVPR, wang2019deep, zhou2019relation} mainly focus on feature enhancement based on instance regions.

More recently, it has become popular to treat object detection as a keypoint estimation task \cite{law2018cornernet,zhou2019objects} to improve detection efficiency. Following the same idea, IPNet \cite{Wang2020IPNet} and PPDM \cite{liao2019ppdm} pose HOI detection as a keypoint detection and grouping problem, which propose to build end-to-end HOI detectors. Specifically, PPDM \cite{liao2019ppdm} contains two parallel branches to predict three points (i.e., human, object and interaction points) and two displacements from the interaction point to its corresponding human and object points. Then, human with its corresponding object can be paired using the interaction point and displacements as bridge. However, without relation reasoning, these end-to-end HOI detectors lack learning HOI-specific interactive semantics in their methods, thus may be deficient in modelling HOIs (e.g., bring problem of miss-pairing between human and object). 
%So there will be problems such as mismatch between human and object.

\begin{figure*}
\begin{center}
%\fbox{\rule{0pt}{2in} \rule{.9\linewidth}{0pt}}
\includegraphics[width=1\linewidth]{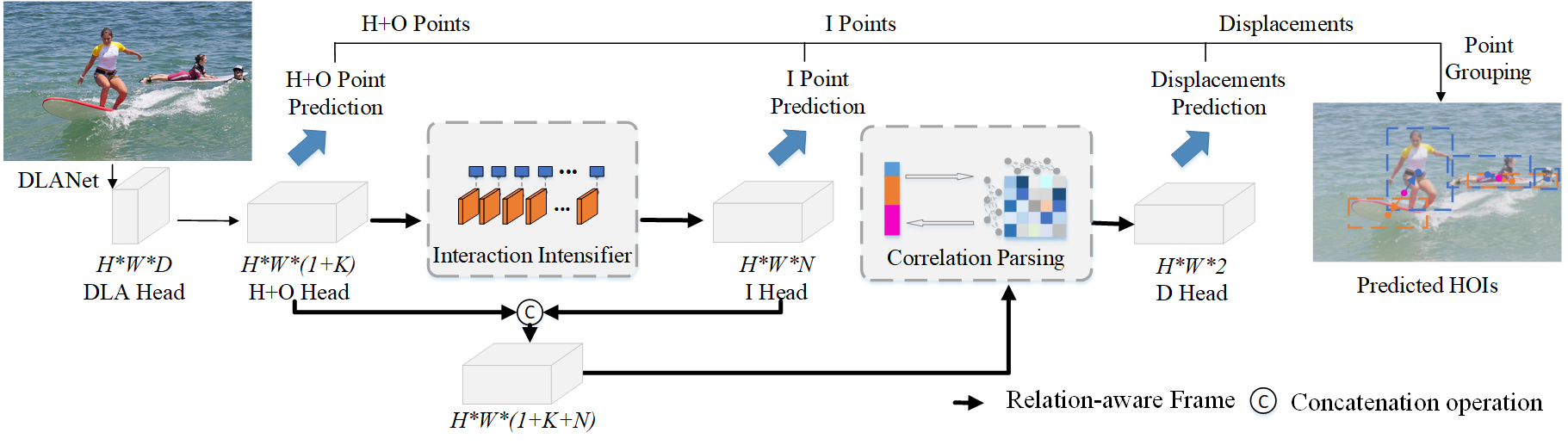}
\end{center}
\vspace{-0.7cm}
   \caption{An overview of RR-Net. The framework is coherent, where instance points prediction, interaction points prediction and displacements prediction are treated as highly correlated processes. We also develop an Interaction Intensifier Module and a Correlation Parsing Module to achieve efficient relation reasoning to learn region-independent interactive semantics for HOI inference.}
\label{FIG:2}
\vspace{-0.5cm}
\end{figure*}

\subsection{Feature Learning in HOI Detection}
%\paragraph{Contextual Cues.}
\textbf{Contextual Cues.}
Several contextual cues have been explored to improve HOI detection.
%%%, of which spatial configuration and human pose have become the most popular cues. 
Given a $\langle$ human, object $\rangle$ candidate with bounding boxes, the interaction pattern \cite{chao2018learning} or spatial configuration \cite{gao2018ican:,VSGNet_2020_CVPR} is a coarse region layout encoded by a binary image with two channels. Tanmay \cite{No-Frills_2019_ICCV} encodes both absolute and relative position of the human and object boxes. Additionally, human pose obtained from pre-trained pose estimators \cite{fang2017rmpe} is unitized as a valid cue to support HOI detection. With human pose vector, human body-parts \cite{fang2018pairwise,wan2019pose} are explored as a set of local regions centered at human keypoints with certain size. Although extracting region-based spatial encodings or body-parts improve detection results, these methods are not favored since additional computation and premade annotations are indispensable. 

%\paragraph{Appearance Attention and Reasoning.} 
\textbf{Appearance Attention and Reasoning.} 
%\cite{sharma2015action}
Inspired by attention mechanism in action recognition, Contextual Attention \cite{wang2019deep} and GID-Net \cite{YANG2020GID} propose to capture instance-centric attention features and context-aware appearance features for human and object. PMFNet \cite{wan2019pose} focuses on pose-aware attention by employing attentional human parts. VSGNet \cite{VSGNet_2020_CVPR} proposes to refine visual features by the spatial configuration of human-object pair. Moreover, GPNN \cite{qi2018learning} and in-GraphNet \cite{in_GraphNet_ijcai2020} introduce learnable graph-based structures, in which HOIs are represented with graphs. RPNN \cite{zhou2019relation} introduces an object-body graph and a human-body graph to capture contexts between body-parts and surrounding instances, where fine-grained human body-part regions are required as prior information.

Overall, methods employing appearance attention mechanisms or graphs are enlightening but complicated, e.g., implementing attention mechanisms upon $M$ human regions and $N$ object regions needs $M+N$ standalone calculations. Analogously, constructing graphs with $M*N$ possible $\langle$ human, object $\rangle$ pairs needs to calculate $M*N$ candidate samples. More importantly, above methods neglect relation reasoning to explore implicit interactive semantics among humans, objects and interactions, whereas comprehending relation is essential to improve HOI detection.
In this paper, unlike the previous techniques of HOI feature learning, we bring region-independent relation reasoning to integrate and transmit HOI-specific interactive semantics for HOI inference.
\begin{figure}
\begin{center}
%\fbox{\rule{0pt}{2in} \rule{.9\linewidth}{0pt}}
\includegraphics[width=1\linewidth]{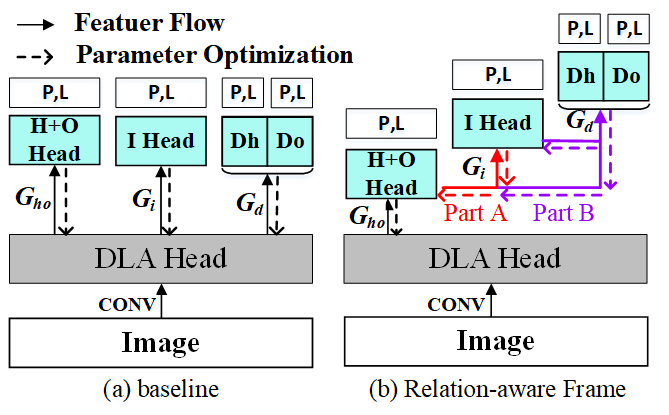}
\end{center}
\vspace{-0.5cm}
   \caption {Relation-aware Frame visualization. (a) In baseline, different heads are built independently, and head parameters are optimized severally in backward propagation. (b) Relation-aware Frame: building I head upon H and O heads. Meanwhile, building two D heads upon all points heads (i.e., H, O and I heads). Here, P means prediction operation and L means loss function.}
\label{FIG:3}
\vspace{-0.5cm}
\end{figure}

\section{Proposed Method}

\subsection{Overview of RR-Net}
An overview of our proposed RR-Net is shown in Figure~\ref{FIG:2}. DLA-34 \cite{Yu2018DLA} is employed as the backbone network in our implementation to extract the global $128*128*64$ DLA feature from an image. Our framework formulates HOI detection as a point detection and grouping task \cite{Wang2020IPNet,liao2019ppdm}, where predicting HOI triplets consists of three steps: 1) predicting human points (H points), $K$-class object points (O points) and $N$-class interaction points (I points). Here, I point is defined as the midpoint of a corresponding $\langle$ human, object $\rangle$ pair. 2) predicting two displacements from the I point to the H point and O point respectively, where each displacement contains an abscissa value and an ordinate value. 3) grouping each I point with the H point and O point respectively to generate a set of triplets based on the predicted points and displacements.
%%\emph{${dp}=({dp}_x, {dp}_y)$}

We have noticed that most methods focus on HOI feature learning but none of them study the association between HOI cognitive processes or explore interactive semantics in these processes. However, recognizing HOIs is a holistic task, in which each process is closely related to the others and provides significant interactive semantics for HOI detection. In order to explore above information, a Relation-aware Frame (See Section 3.2) is first proposed to provide a coherent structure and parameter sharing pattern for interaction inference.
%%%Specifically, we consider points and displacements prediction as highly correlated processes, so that we link all prediction procedures together and optimize their parameters connectedly.
%%For instance, given H and O feature heads obtained from DLA head, we inherit H and O heads to build I head, rather than generating I head upon DLA head in parallel. Meanwhile, displacement feature head (D head) is built upon all point heads.
Upon the frame, an Interaction Intensifier Module (See Section 3.3) and a Correlation Parsing Module (See Section 3.4) are then developed to achieve region-independent relation reasoning so that interactive semantics and correlations among visual targets can be efficiently integrated. Next, we will introduce the proposed framework in detail.

\subsection{Relation-aware Frame}%\lable{sec:series} (See Section ~\ref{sec:rim})
As illustrated in Figure~\ref{FIG:3} (a), to predict H points, O points, I points and two displacements, the baseline build different prediction feature heads independently by the function group \emph{$G(·)=[G_{ho}(·), G_i(·), G_{dh}(·), G_{do}(·)]$}, and head parameters are optimized severally in backpropagation. It should be mentioned that we have merged $G_{dh}(·)$ and $G_{do}(·)$ as $G_{d} (·)$ in Figure~\ref{FIG:3} for simplicity. We argue that such a structure neglects the implicit relation among these predictions. It is noted that the human visual system usually detects instances and groups related ones coherently from the scene to recognize a HOI. Considering the triplet $\langle$ human, ride, bike $\rangle$ as an example, the person and bike may be detected firstly from the scene. Then, semantics obtained during above process are not discarded but further used to determine the category and elements of interaction. Based on the above inspiration, we propose a correlated and coherent structure for interaction inference, which is termed as a Relation-aware Frame.

For predicting humans and objects, we first generate H and O head feature \emph{$f_{ho}\in\Re^{H*W*(1+K)}$} from global DLA head feature \emph{$f_{dla}\in\Re^{H*W*D}$} by the function \emph{$G_{ho}(·)$}, where $H*W$ denotes range, $D$ denotes feature dimension and $(1+K)$ is the category synthesis of human and $K$-class objects. 
The two stages of our Relation-aware Frame are then established to break the independent predictions of baseline, which are: 1) Part A: building I head feature \emph{$f_i\in\Re^{H*W*N}$} upon feature $f_{ho}$ by the function \emph{$G_i(·)$}; 2) Part B: building two D head features \emph{$f_{dh}, f_{do}\in\Re^{H*W*2}$} upon all point head features \emph{$ f_p=concat(f_{ho}, f_i)$} by the function \emph{$G_{d}(·)$}. 
%%For ease of understanding, we provide Figure~\ref{FIG:3} (b) and (c) to show these two stages respectively. 
As illustrated in Figure~\ref{FIG:3} (b), in the fully Relation-aware Frame, output head features of each previous stage are used as input in the next stage.
The function group \emph{$G(·)$} are computed by:
\begin{equation}
G(X)= Conv2d(ReLU(Conv2d(X))),
\label{equ1}
\end{equation}
where the first $3*3$ convolution transforms the features from input dimension to hidden dimension, and the second $1*1$ convolution acts as a fully connected layer to reduce the feature from hidden dimension to specified dimension to match the predictions. The next predictions of H points, O points and I points can be regarded as heatmap estimation processes by sigmoid functions. In this way, instance points prediction loss \emph{$L_{ho}$}, interaction point prediction loss \emph{$L_i$} and two displacements prediction loss \emph{$L_d=L_{dh} + L_{do}$} are correlated in backpropagation and parameters are optimized connectedly.
%%According to our progressive Relation-aware Frame, the partial derivative of each prediction loss to DLA head weight is calculated as:

% \begin{small}
% \begin{equation}
% \frac{\partial L_{ho}}{\partial W_{dla}} = \frac{\partial L_{ho}}{\partial W_{ho}F_{ho}} * \frac{\partial W_{ho}F_{ho}}{\partial W_{dla}} ,
% \label{equ2}
% \end{equation}
% \end{small}
% \begin{small}
% \begin{equation}
% \frac{\partial L_{i}}{\partial W_{dla}} = \frac{\partial L_{i}}{\partial W_{i}F_{i}} * \frac{\partial W_{i}F_{i}}{\partial W_{ho}F_{ho}} * \frac{\partial W_{ho}F_{ho}}{\partial W_{dla}} ,
% \label{equ3}
% \end{equation}
% \end{small}
% \begin{small}
% \begin{equation}
%   \begin{split}
%   \frac{\partial L_{dh}}{\partial W_{dla}} = \frac{\partial L_{dh}}{\partial W_{dh}F_{dh}} * (\frac{\partial W_{dh}F_{dh}}{\partial W_{i}F_{i}} * \frac{\partial W_{i}F_{i}}{\partial W_{ho}F_{ho}} * \\ \frac{\partial W_{ho}F_{ho}}{\partial W_{dla}} + \frac{\partial W_{dh}F_{dh}}{\partial W_{ho}F_{ho}} * \frac{\partial W_{ho}F_{ho}}{\partial W_{dla}}) , 
%   \end{split}
% \label{equ4}
% \end{equation}
% \end{small}
% \begin{small}
% \begin{equation}
%   \begin{split}
%   \frac{\partial L_{do}}{\partial W_{dla}} = \frac{\partial L_{do}}{\partial W_{do}F_{do}} * (\frac{\partial W_{do}F_{do}}{\partial W_{i}F_{i}} * \frac{\partial W_{i}F_{i}}{\partial W_{ho}F_{ho}} * \\ \frac{\partial W_{ho}F_{ho}}{\partial W_{dla}} + \frac{\partial W_{do}F_{do}}{\partial W_{ho}F_{ho}} * \frac{\partial W_{ho}F_{ho}}{\partial W_{dla}}) .   
%   \end{split}
% \label{equ5}
% \end{equation}
% \end{small}

\begin{figure}
\begin{center}
%\fbox{\rule{0pt}{2in} \rule{.9\linewidth}{0pt}}
\includegraphics[width=1\linewidth]{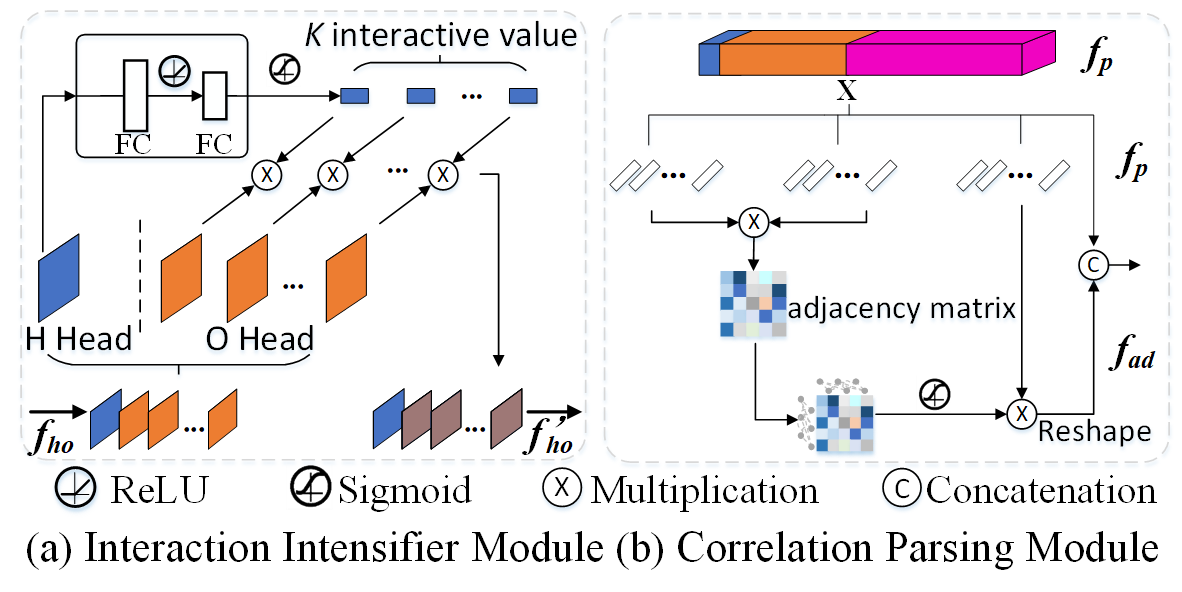}
\end{center}
\vspace{-0.5cm}
   \caption {The detailed design of proposed IIM and CPM. IIM takes H and O head features as inputs and passes interactive semantics from humans to objects to intensify interactions. CPM exploits and integrates interactive correlations among visual targets and helps group HOI triplets without region proposals.}
\label{FIG:4}
\vspace{-0.5cm}
\end{figure}

\subsection{Interaction Intensifier Module}
Human contexts typically have strong effects to HOIs that different postures and distribution of people often imply information of taking interactions with different objects. For example, we can clearly distinguish between kickers and riders by their postures and distribution. Meanwhile, kickers can easily remind us of football, but riders remind us of bicycles or motorcycles. To exploit interactive semantics above to intensify interactions, we propose an Interaction Intensifier Module (abbr. IIM) upon the Relation-aware Frame Part A.
%To utilize implicit relation above, we propose an Interaction Intensifier Module (abbr. IIM), where interactive semantics can be exploited and passed from humans to objects but without any region proposals. 

After running $G_{ho}(·)$, each obtained feature map of H and O heads contains characteristic information for prediction of corresponding instances. We innovatively employ these features to perform relation reasoning, where no human and object regions are needed. 
As mentioned above, the total number $(1+K)$ of categories contains person and $K$ other object categories. As illustrated in Figure~\ref{FIG:4} (a), We can further split  \emph{$f_{ho}\in\Re^{H*W*(1+K)}$} into \emph{$f_h\in\Re^{H*W*1}$} and \emph{$f_o\in\Re^{H*W*K}$}. Then, a multilayer perceptron (MLP) with two fully connected layers followed by a sigmoid operation is adopted on feature \emph{$f_h$} to generate the embedded interactive vector \emph{$\beta = [\beta_1,\beta_2,...,\beta_K]$}, denoting relevance between humans and different objects. 
Afterwards, O head features \emph{$f_o$} are weighted with vector \emph{$\beta$} by a multiplication operation, generating the valued O head features \emph{$f'_o$}.
\begin{equation}
\beta = sigmoid(MLP(f_h)), f'_o = \beta * f_o,
\label{equ6}
\end{equation}
where \emph{$\beta_k\in[0,1]$} is the $k$-th interactive value for $f_o$. Finally, IIM outputs \emph{$f'_{ho}=concat(f_h, f'_o)$}, which is in same form as inputs.
Differing from treating human and object features separately, with the formulation above, interactive semantics from humans can be dynamically passed to objects to intensify  interactions and promote I points prediction. 

\subsection{Correlation Parsing Module}
Interactive correlations among humans, objects, and interactions provide crucial information for HOI inference which are different from region-based instance representations. We propose Correlation Parsing Module (abbr. CPM) upon the Relation-aware Frame Part B to break away from the previous idea of region-based feature enhancement and exploit the interactive correlations among these targets. The detailed design of CPM is provided in Figure~\ref{FIG:4} (b).
The CPM takes all point head features \emph{$f_p\in\Re^{H*W*(1+K+N)}$} as input, where the human class, $K$ object classes and $N$ interaction classes are considered as $(1+K+N)$ visual targets. 
%% We first propose a project function to map all point head feature \emph{$f_p$} into a (1+K+N)-wide adjacency matrix. A message spreading process is then adopted to parse interactive correlations among visual targets. Furthermore, the updated matrix is reverse projected to feature \emph{$f_{ad}$} with correlative information. Finally, \emph{$f_p$} and \emph{$f_{ad}$} are concatenated to provide both appearance and correlative information for the next displacements prediction. 

In particular, a project function first provides a pattern to convert head feature \emph{$X=f_p$} into a dynamic adjacency matrix $A$.
%%%The calculation can be divided into two parts: feature embedings and a linear combination. 
Specifically, given the input \emph{$X\in\Re^{H*W* (1+K+N)}$}, we embed two feature tensors \emph{$(X^\theta, X^\phi)$} by \emph{$\theta(X)$} and \emph{$\phi(X)$} modeled by two $1*1$ convolutions. The obtained tensors are then reshaped from \emph{$H*W*(1+K+N)$} to \emph{$L*(1+K+N)$} by the {$Planar$} operation, where their two-dimensional location pixels of \emph{$H*W$} are converted to one-dimensional vector \emph{$L$}. Finally, a linear combination transforms visual targets into the unified adjacency matrix \emph{$A\in\Re^{(1+K+N)* (1+K+N)}$}, where \emph{$(1+K+N)$} denotes the length and width of the adjacency matrix. Each element \emph{$a_{ij}\in{A}$} denotes the correlation between two visual targets. 
\begin{equation}
A = Planar[\theta(X)]^T * Planar[\phi(X)].
\label{equ8}
\end{equation}

Based on the adjacency matrix, a message spreading process is further adopted to broadcast and integrate interactive correlations from all elements, which can be computed by:
\begin{equation}
A' = Conv1D((Conv1D(A)^T)^T\oplus A),
\label{equ10}
\end{equation}
where two operations of \emph{$Conv1D$} are single-layer 1D convolutions to build communication among all elements. Between the convolutions, the \emph{$\oplus$} implements addition point to point which updates the hidden element states according to incoming information. The \emph{$A'$} is the updated adjacency matrix.
%that perform Laplacian smoothing

The following step provides a reverse projection of $A'$ to convolution space.
%, which outputs a feature tensor $f_{ad}$. 
Firstly, a \emph{$g (X)$} function and a reshape operation act as linear embedding to compute the representation of input signal as visual targets. Then a weighted linear combination is employed as the reverse projection:
\begin{equation}
Y = sigmoid(A') * Planar[g(X)].
\label{equ11}
\end{equation}
In this way, \emph{$A'$} plays a role of dynamical weights so that interactive correlations among visual targets can be weighted aggregated. Finally, we reshape the planar feature tensor \emph{$Y\in\Re^{L*(1+K+N)}$} to three-dimensional \emph{$f_{ad}\in\Re^{H*W*(1+K+N)}$}. 
%The proposed CPM is light since all parameters are end-to-end learnable and come from 1*1 convolutions. 
For the next displacements prediction, \emph{$f_{ad}$} as the correlation encoding is concatenated with appearance feature \emph{$f_p$} to provide correlative information among all visual targets.

\subsection{Training and Inference}
%\paragraph {Training.}
\textbf {Training.}
HOI detection needs to generate not only the category of instances and interactions, but also the bounding boxes of instances. Therefore, we also need to regress the size and offset of instance boxes by \emph{$G_{wh}(·)$} and \emph{$G_{off}(·)$}, which have the same computation with {$G(·)$} in Section 3.2. In summary, the final loss {$L = L_p + \lambda L_d + L_r$} can be calculated as the weighted sum of point prediction loss \emph{$L_p$}, displacement prediction loss $L_d$ and instance box regression loss $L_r$,
%\begin{equation}
%L = L_p + \lambda L_d + L_r,
%\label{equ12}
%\end{equation}
where $\lambda$ is set as 0.1 \cite{zhou2019objects}.
As mentioned in Section 3.2, three heatmaps can be produced by employing sigmoid upon the H, O and I head features respectively. We also map ground-truth H, O, and I points into three heatmaps with a Gaussian kernel respectively. Therefore, point detection can be transformed into a heatmap estimation problem, where \emph{$L_p=L_{ho}+L_i$} are optimized by element-wise focal loss. 
The \emph{$L_d=L_{dh}+L_{do}$} is trained by $L1$ loss, where $L_{dh}$ and $L_{do}$ are losses for predicted displacement \emph{${dp}^{ih}$} to H point and displacement \emph{${dp}^{io}$} to O point respectively for an I point.
The \emph{$L_r=L_{wh}+L_{off}$} is optimized by $L1$ loss, which is in charge of regressing box size \emph{$(h, w)$} and local offset \emph{$d=(d_x, d_y)$} for location of ground truth instance points. 
%%%%More details can be found in the supplementary materials.
%%\emph{${dp}^{ih}=({dp}_x^{ih}, {dp}_y^{ih})$}
%%\emph{${dp}^{io}=({dp}_x^{io}, {dp}_y^{io})$}

\begin{table}
\centering
\resizebox{82mm}{!}{
\begin{tabular}{llc}
\toprule
Method & Backbone Network & mAP$_{role}$(\%)   \\ \hline
%%%%Gupta et al.\cite{yatskar2016situation}  & ResNet-50-FPN & 31.8  \\
InteractNet \cite{gkioxari2018detecting}  & ResNet-50-FPN & 40.0  \\
%iHOI \cite{xu2019interact}  & ResNet-50 & 40.4  \\
GPNN \cite{qi2018learning}  & Deformable CNN & 44.0  \\
iCAN \cite{gao2018ican:}   & ResNet-50 & 45.3  \\
Contextual Att \cite{wang2019deep}   & ResNet-50 & 47.3 \\
RPNN \cite{zhou2019relation}   & ResNet-50 & 47.5  \\
TIN(R$P_dC_d$) \cite{li2019transferable}   & ResNet-50 & 47.8  \\
in-GraphNet \cite{in_GraphNet_ijcai2020} & ResNet-50 & 48.9 \\
IPNet \cite{Wang2020IPNet} & Hourglass-104 & 51.0  \\
PMFNet \cite{wan2019pose} & ResNet-50 & \underline{52.0}  \\ \hline
our baseline   & DLA-34 & 51.4  \\
\textbf{RR-Net} & DLA-34 & \textbf{54.2}  \\
\bottomrule
\end{tabular}}
\caption{Performance comparison on V-COCO.}
\label{table1}
\vspace{-0.5cm}
\end{table}

%\paragraph{Inference.}
\textbf {Inference.}
After points detection, we select top $T$ human, object and interaction points according to the corresponding confidence scores respectively. Then, point grouping process will determine the optimal triplets. Specifically, for each detected I point \emph{$(x^i, y^i)$} and predicted displacements  \emph{${dp}^{ih}$} and \emph{${dp}^{io}$}, several candidate instance points will be ranked according to two criteria, which are 1) have high confidence score, 2) be close to the human/object location that generated by interaction point location plus the human/object displacement. %%%%More information can be found in the supplementary materials.
%%predicted displacements \emph{$({dp}_x^{ih}, {dp}_y^{ih})$} and \emph{$( {dp}_x^{io}, {dp}_y^{io})$}

For a grouped regressive H point \emph{$(x^h, y^h)$} and an O point \emph{$(x^o, y^o)$}, their bounding boxes are determined as \emph{$B_h=(x^h,y^h,h^h,w^h)$} and \emph{$B_o=(x^o,y^o,h^o,w^o)$},
where $(h^h,w^h)$ and $(h^o,w^o)$ denote the width and height of the human box and object box, respectively. With a set of detected HOI triplets, we use a fusion of scores to predict the final score $S_{hoi}$ for each of them:
\begin{equation}
S_{hoi}=S_{(x^h,y^h)}^h * S_{(x^i,y^i )}^i * S_{(x^o,y^o)}^o,
\label{equ19}
\end{equation}
where \emph{$S_{(x^h,y^h)}^h$}, \emph{$S_{(x^i,y^i)}^i$} and \emph{$S_{(x^o,y^o)}^o$} are confidence scores of candidate H point, I point and O point.

\begin{table}
\centering
\resizebox{82mm}{!}{
\begin{tabular}{llccc}
\toprule
\multirow{2}{*}{Method} & \multirow{2}{*}{Backbone Network} & \multicolumn{3}{c}{Default} \\ \cline{3-5}
 & & \multicolumn{1}{c}{full} & \multicolumn{1}{c}{rare} & \multicolumn{1}{c}{non-rare} \\ \hline
%%%%Shen et al. \cite{shen2018scaling}  & VGG-19 &6.46 &4.24 &7.12  \\
%%%%HO-RCNN \cite{chao2018learning}  & CaffeNet & 7.81 & 5.37 & 8.54    \\
InteractNet \cite{gkioxari2018detecting}  &ResNet-50-FPN &9.94 &7.16 &10.77    \\
%iHOI \cite{xu2019interact}  &ResNet-50 & 9.97 & 7.11  &10.83  \\
GPNN \cite{qi2018learning}  &Deformable CNN & 13.11 & 9.34 & 14.23    \\
iCAN \cite{gao2018ican:}  & ResNet-50 & 14.84 & 10.45 & 16.15     \\
Contextual Att \cite{wang2019deep}  & ResNet-50 & 16.24 & 11.16 & 17.75     \\ 
RPNN \cite{zhou2019relation}  &ResNet-50 & 17.35 & 12.78  & 18.71    \\
TIN(R$P_dC_d$) \cite{li2019transferable}  &ResNet-50 & 17.03 & \underline{13.42} & 18.11     \\ 
No-Frills \cite{No-Frills_2019_ICCV}  &ResNet-50 & 17.18 & 12.17 & 18.68     \\ 
PMFNet \cite{wan2019pose}  &ResNet-50 & 17.46 & \textbf{15.65} & 18.00     \\ 
%%%PPDM \cite{liao2019ppdm}  &DLA-34 & 19.02 & 12.65 & 20.92     \\ 
IPNet \cite{Wang2020IPNet}  &Hourglass-104 & 19.56 & 12.79 & 21.58     \\ 
PPDM \cite{liao2019ppdm}  &DLA-34 & \underline{20.29} & 13.06 & \underline{22.45}     \\ \hline
our baseline  & DLA-34 & 18.87 & 11.32  & 21.13   \\
%%%\textbf{RR-Net}  & DLA-34 & \textbf{19.80} & 11.48 & \textbf{22.29}     \\
\textbf{RR-Net}  & DLA-34 & \textbf{20.72} & 13.21 & \textbf{22.97}     \\
\bottomrule
\end{tabular}}
\caption{Performance comparison on HICO-DET.}
% test set. Default: all images. Full: all 600 HOI categories. Rare: 138 HOI categories with less than 10 training instances. Non-Rare: 462 HOI categories with 10 or more training instances.}
\label{table2}
\vspace{-0.5cm}
\end{table}

\begin{table}
\centering
\resizebox{75mm}{!}{
\begin{tabular}{lc}
\toprule
Method                             & mAP$_{role}$(\%) \\ \hline
baseline                           & 51.4 \\
baseline+Relation-aware Frame Part A        & 52.4 \\
baseline+Relation-aware Frame Part B        & 52.2 \\
baseline+Fully Relation-aware Frame        & 52.8 \\
\bottomrule
\end{tabular}}
\caption{ Impact of adopting progressive Relation-aware Frame.}
\label{table3}
\vspace{-0.3cm}
\end{table}

\begin{table}
\centering
\resizebox{82mm}{!}{
\begin{tabular}{lcccccc}
\toprule
Baseline            &{$\surd$} &  & &  &  &    \\
Relation-aware Frame Part A  & &{$\surd$} & & {$\surd$} & {$\surd$}&  {$\surd$}     \\
Relation-aware Frame Part B  & & &{$\surd$} &{$\surd$} & {$\surd$} & {$\surd$}    \\   \hline
IIM                 & &{$\surd$} & & {$\surd$}&  & {$\surd$}    \\   \hline
CPM                 & & &{$\surd$} & & {$\surd$} & {$\surd$}    \\   \hline
mAP$_{role}$(\%) & 51.4 & 52.9 & 52.8 & 53.6 & 53.4 & 54.2 \\
\bottomrule
\end{tabular}}
\caption{ Impact of employing IIM and CPM.}
\label{table4}
\vspace{-0.5cm}
\end{table}

\section{Experiments and Evaluations}

\subsection{Experimental Setup}
%\paragraph{Datasets and Evaluation Metrics.} 
\textbf{Datasets and Evaluation Metrics.} 
We evaluate our method on two large-scale benchmarks, including V-COCO \cite{yatskar2016situation} and HICO-DET \cite{chao2018learning} datasets. V-COCO includes 10,346 images, which contains 16,199 human instances in total and provides 26 common verb categories. HICO-DET contains 47,776 images, where 80 object categories and 117 verb categories compose of 600 HOI categories. There are three different HOI category sets in HICO-DET, which are: (a) all 600 HOI categories (Full), (b) 138 HOI categories with less than 10 training instances (Rare), and (c) 462 HOI categories with 10 or more training instances (Non-Rare). 
A detected $\langle$ human, verb, object $\rangle$ triplet is considered as a true positive if: 1) it has correct interaction label; 2) both the predicted human and object bounding boxes have IoU of 0.5 or higher with the ground-truth boxes. Following the standard protocols, we use role mean average precision $(mAP_{role})$ \cite{yatskar2016situation} to report evaluation results.

\begin{figure}[t]
\begin{center}
\includegraphics[width=1\linewidth]{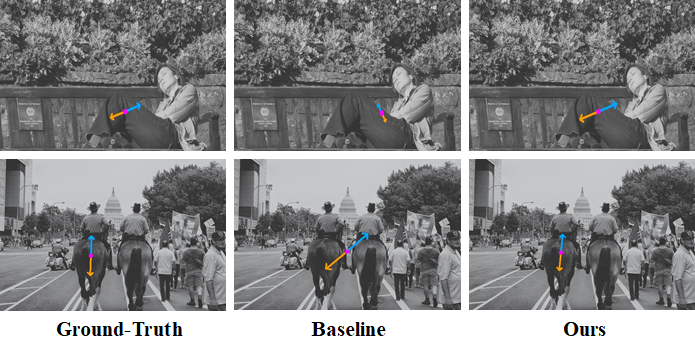}
\end{center}
\vspace{-0.5cm}
   \caption{Visualized effects of IIM and CPM. The first column shows examples of ground-truth annotation. The second column visualizes interaction point and displacements predictions before point grouping using baseline method. The third column represents above predictions with IIM and CPM. Best viewed in color.}
\label{FIG:5}
\vspace{-0.5cm}
\end{figure}

\begin{figure}
\begin{center}
%\fbox{\rule{0pt}{2in} \rule{.9\linewidth}{0pt}}
\includegraphics[width=1\linewidth]{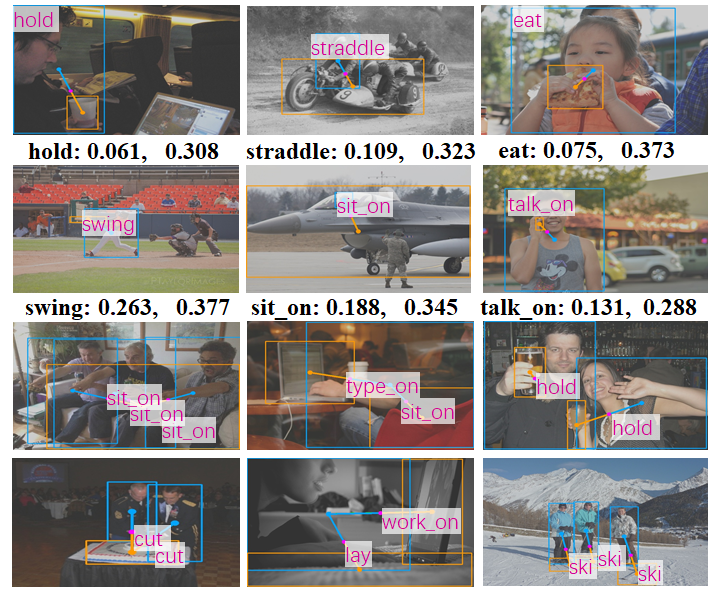}
\end{center}
\vspace{-0.5cm}
   \caption {Visualization of HOI detections. The first two rows show our results compared with baseline, where texts below indicate the detected interaction (verb), and two numbers in turn represent scores predicted by baseline and our method. The third and fourth rows show multiple people take interactions with various objects concurrently detected by our method. Best viewed in color.}
\label{FIG:6}
\vspace{-0.5cm}
\end{figure}

%\paragraph{Implementation details.} 
\textbf{Implementation details.} 
During training, input images have the resolution of $512*512$, yielding a resolution of $128*128$ for all output head features. We employ standard data augmentation following \cite{zhou2019objects}. Our model is optimized with Adam. The batch-size is set as 15 for V-COCO and 20 for HICO-DET. We train the model (on a single GPU) for 140 epochs, with the initial learning rate of 5e-4 which drops 10x at 90 and 120 epochs respectively. The top predictions $T$ is set as 100. All our experiments are conducted by Pytorch on a GPU of NVIDIA Tesla P100.

\subsection{Overall Performance}
We compare our method with state-of-the-arts in this subsection. Meanwhile, we strip all components related to the Relation-aware Frame structure, IIM and CPM as baseline. 
%%%Interaction Intensifier Module and Correlation Parsing Module

%\paragraph{ Performance on V-COCO.}
\textbf{ Performance on V-COCO.}
Comparison results on V-COCO in terms of \emph{$mAP_{role}$} are shown in Table~\ref{table1}. It can be seen that our proposed RR-Net has a {mAP(\%)} of 54.2, obtaining the best performance among all methods. Although we do not adopt previous region-based feature learning (e.g., RPNN \cite{zhou2019relation}, Contextual Att \cite{wang2019deep}), or employ additional human pose (e.g., PMFNet \cite{wan2019pose}, TIN \cite{li2019transferable}), our method outperforms these methods with sizable gains. Besides, our method achieves an absolute gain of 2.8 points, which is a relative improvement of 5.5\% compared with the baseline, validating its efficacy in HOI detection task. 

%\paragraph{ Performance on HICO-DET}
\textbf{Performance on HICO-DET.}
Table~\ref{table2} shows the comparisons of RR-Net with state-of-the-arts on HICO-DET. Firstly, the detection results of our RR-Net are the best among all methods under the Full and Non-Rare settings, demonstrating that our method is more competitive than the others in detecting most common HOIs. It is noted that RR-Net is not preeminent in detecting rare HOIs (HOI categories with less than 10 training instances), because our relation reasoning modules need enough training samples to exert their effects. Besides, our RR-Net obtains 20.72 mAP on HICO-DET (Defualt Full), which achieves relative gain of 9.8\% compared with the baseline. These results quantitatively show the efficacy of our method.
%%% absolute gain of 0.9 point and 

\subsection{Ablation Studies}
%We conduct several ablation studies in this subsection. V-COCO \cite{yatskar2016situation} serves as the testbed on which we further analyze the individual effect of components in our method.
%\paragraph{ Performance Impact of Proposed Components }
\textbf{ Performance Impact of Proposed Components.}
V-COCO \cite{yatskar2016situation} serves as the testbed in this subsection. We first study the performance impact of our proposed progressive Relation-aware Frame, Interaction Intensifier Module (IIM) and Correlation Parsing Module (CPM). As shown in Table~\ref{table3}, when we adopt the Relation-aware Frame Part A or Part B (See Figure~\ref{FIG:3}) severally, the mAP are 52.4 and 52.2, indicating the respective effects of the two parts. Adopting the fully Relation-aware Frame obtains the mAP of 52.8, which boosts the performance by 1.4 points compared with the baseline. Meanwhile, we investigate the impact of employing IIM and CPM in Table~\ref{table4}. Compared with the baseline, employing IIM (upon Relation-aware Frame Part A) and CPM (upon Relation-aware Frame Part B) brings gains of 1.5 mAP and 1.4 mAP, respectively. Furthermore, improvements of 2.2 mAP and 2.0 mAP can be obtained while employing IIM and CPM respectively upon our fully Relation-aware Frame. Finally, RR-Net achieves an absolute gain of 2.8 mAP compared with the baseline. These evidences show that each proposed component of our region-independent relation reasoning indeed contributes to the final performance. 

\textbf{Effects visualization of IIM and CPM.}
Moreover, we take several examples of interaction point and displacements predictions before point grouping in Figure~\ref{FIG:5} to visualize the effects of IIM and CPM. Intuitively, with IIM and CPM, our framework makes better predictions of interaction point location and displacements compared with baseline, which verifies the necessity of relation reasoning. It is worth noting that due to the inaccurate prediction of interaction point location and displacements, the baseline framework in the second row may lead to miss-pairing between the human and horse. However, our framework provides predictions similar to the ground-truth annotation, which helps produce correct HOI detection in point grouping procedure. 

\subsection{Qualitative Examples}
%\textbf{Qualitative Examples}
In Figure~\ref{FIG:6}, we first compare our results with baseline in the first two rows to demonstrate our improvements. Each subplot displays one detected $\langle$ human, verb, object $\rangle$ triplet for easy observation.
%%%%, including the location of person and object, as well as the interaction between the above two instances. 
We can see that our method is capable of detecting various HOIs with higher scores. In addition, the third and fourth rows show that our method is able to detect multiple people taking different interactions with diversified objects. Summarizing all above results validates that our method can learn HOI-specific semantics well from complex environments to provide high quality HOI detections.

\section{Conclusion}
Recent end-to-end HOI detectors are limited in modelling HOIs due to they are incapable to perform relation reasoning in their methods. Besides, previous region-based HOI feature learning is hardly applicable to end-to-end detectors. Therefore, we make a first try to develop a RR-Net to explore novel techniques of region-independent relation reasoning, which opens up a new direction to improve HOI detection.
Specifically, a Relation-aware Frame is proposed, upon which an Interaction Intensifier Module and a Correlation Parsing Module are designed successively to reason HOI-specific interactive semantics without instance regions.
Extensive experiments show that our RR-Net outperforms existing methods with sizable gaps, validating its efficacy in detecting HOIs. 

\bibliographystyle{named}
\bibliography{ijcai21}

\end{document}